\RequirePackage{fix-cm}

\documentclass[twocolumn]{svjour3}          
\smartqed  
\usepackage{subcaption}
\usepackage{graphicx}
\usepackage[export]{adjustbox}
\usepackage[ruled,vlined]{algorithm2e}
\usepackage{multirow}
\usepackage[noend]{algpseudocode}
\usepackage{subcaption}
\usepackage{xcolor}
\usepackage{hyperref}
\usepackage{comment}
\usepackage{multicol}
\usepackage{xcolor}
\usepackage{MnSymbol}
\usepackage{float}
\usepackage{tikz}
\begin{document}
	
	\title{MSDNet: Multi-Scale Decoder for Few-Shot Semantic Segmentation via Transformer-Guided Prototyping}

	
	\author{Amirreza Fateh \and Mohammad Reza Mohammadi \and Mohammad Reza Jahed Motlagh}
	

	\institute{Amirreza Fateh \at
		\email{amirreza\_fateh@comp.iust.ac.ir}           
		\and
		Mohammad Reza Mohammadi \at
		corresponding author \\
		\email{mrmohammadi@iust.ac.ir}
		\and
		Mohammad Reza Jahed Motlagh \at
		\email{jahedmr@iust.ac.ir}}
	\maketitle
	\begin{abstract}
		\textbf{: Few-shot Semantic Segmentation addresses the challenge of segmenting objects in query images with only a handful of annotated examples. However, many previous state-of-the-art methods either have to discard intricate local semantic features or suffer from high computational complexity. To address these challenges, we propose a new Few-shot Semantic Segmentation framework based on the Transformer architecture. Our approach introduces the spatial transformer decoder and the contextual mask generation module to improve the relational understanding between support and query images. Moreover, we introduce a multi scale decoder to refine the segmentation mask by incorporating features from different resolutions in a hierarchical manner. Additionally, our approach integrates global features from intermediate encoder stages to improve contextual understanding, while maintaining a lightweight structure to reduce complexity. {This balance between performance and efficiency enables our method to achieve competitive results on benchmark datasets such as $PASCAL\text{-}5^i$ and $COCO\text{-}20^i$ in both 1-shot and 5-shot settings.} Notably, our model with only 1.5 million parameters demonstrates competitive performance while overcoming limitations of existing methodologies. \color{purple}{https://github.com/amirrezafateh/MSDNet}}
		
		
	\end{abstract}

	\section{Introduction}
	\label{intro}
	
	\begin{figure}
		\centering
		\begin{subfigure}[b]{0.95\columnwidth}
			\centering
			\includegraphics[width=\linewidth]{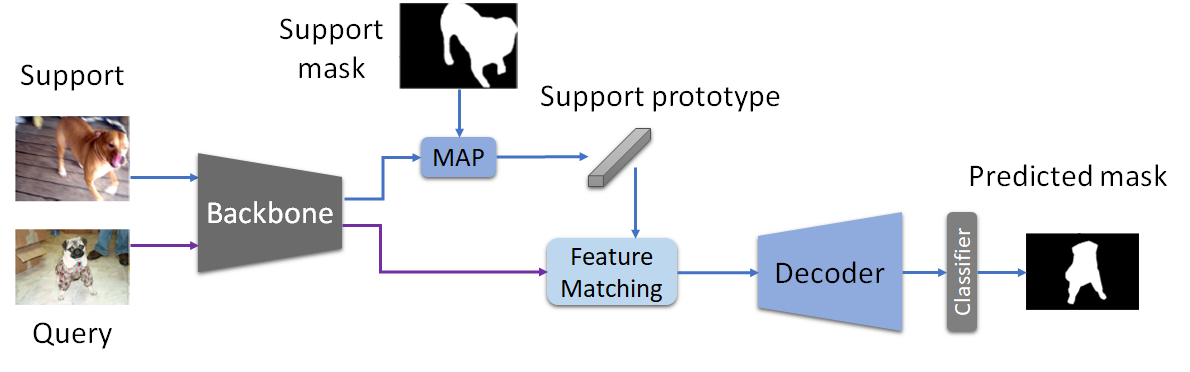}
			\caption{Prototype-based}
			\label{fig:1_a}
		\end{subfigure}
		\newline

		\begin{subfigure}[b]{0.95\columnwidth}
			\centering
			\includegraphics[width=\linewidth]{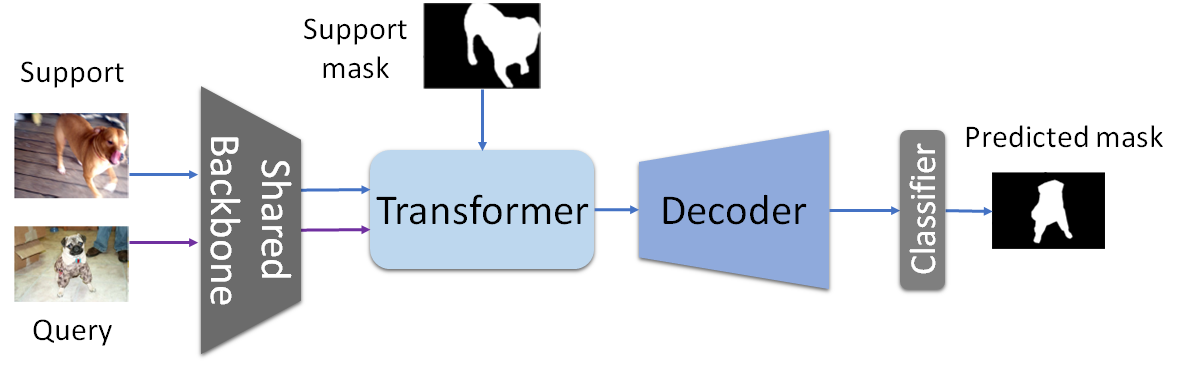}
			\caption{Pixel-wise}
			\label{fig:1_b}
		\end{subfigure}
		\newline

		\begin{subfigure}[b]{0.95\columnwidth}
			\includegraphics[width=\linewidth]{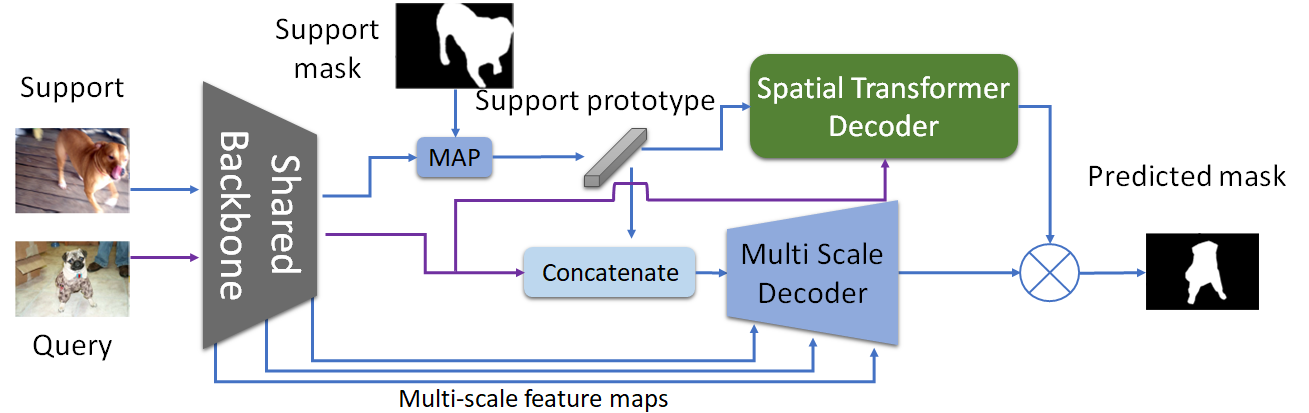}
			\caption{The proposed multi-scale decoder with Transformer-guided prototyping}
			\label{fig:1_c}
		\end{subfigure}

		\caption{Comparison among existing methods and our proposed method for FSS. 
			(a) Prototype-based methods; 
			(b) Pixel-wise methods; 
			(c) The proposed method builds upon prototype-based strategies while enhancing contextual understanding and segmentation quality through Transformer-guided prototyping and multi-scale decoding.
		}

		\label{fig:ch2_adapte_pro}
	\end{figure}
	
	Semantic segmentation is a key task in computer vision, where each pixel of an image is labeled as part of a specific category. This is important in many areas like autonomous driving, medical imaging, and scene understanding \cite{zhang2024segment}. To perform this task well, models need to learn detailed object boundaries. In recent years, deep Convolutional Neural Networks (CNNs) have made big improvements in this area \cite{sun2024remax}. However, these high-performing models usually need large datasets with lots of labeled examples \cite{bi2024prompt,bi2024agmtr,bi2023not}, which takes a lot of time and effort to create. In real-world scenarios, like in medical imaging or other fields where labeled data is limited, this becomes a big problem. To solve this, Few-shot Semantic Segmentation (FSS) has become a useful approach.

	FSS tries to segment new object classes in images using only a few labeled examples, called support images, that show the target class \cite{askari2025enhancing,mao2022bidirectional,mao2024body}. This method helps reduce the need for large datasets, making it more practical for real-world use. Addressing the challenges of FSS requires handling differences in texture or appearance between the target object in the query image and similar objects depicted in the support examples. Effectively using the relationship between the query image and the support examples is essential in tackling FSS. FSS can be widely categorized into two groups: Prototype-based approaches and Pixel-wise methods. 
	
	Prototype-based approaches involve abstracting semantic features of the target class from support images through a shared backbone network. This process results in feature vectors called class-wise prototypes, which are obtained using techniques such as class-wise average pooling or clustering. These prototypes are then combined with query features through operations like element-wise summation or channel-wise concatenation. The combined features are refined by a decoder module to classify each pixel as either the target class or background \cite{li2024dual,sun2023attentional,liu2023learning,ding2023self} (Figure \ref{fig:ch2_adapte_pro}-\subref{fig:1_a}). Pixel-wise methods take a different approach by focusing directly on pixel-level information rather than compressing it into prototypes. These methods aim to predict the target class for each pixel in the query image by comparing it directly with corresponding pixels in the support images. To achieve this, they establish pixel-to-pixel correlations between the support and query features, which allows the model to find precise matches even when the object’s appearance varies. This process is often enhanced by attention mechanisms, like those found in Transformer models, which help the model focus on important relationships between pixels. By avoiding the need for prototypes, Pixel-wise methods aim to preserve more detailed information, allowing for finer-grained segmentation \cite{xu2023self,kang2023distilling,shi2022dense}. An example of this is illustrated in Figure \ref{fig:ch2_adapte_pro}-\subref{fig:1_b}.

	{While both prototype-based and pixel-wise approaches have demonstrated efficacy in few-shot semantic segmentation, they also exhibit key limitations. Prototype-based methods often compress the semantic features of the support images into a single vector, potentially discarding fine-grained spatial information necessary for accurate segmentation—especially for complex object classes. Pixel-wise methods address this by directly comparing individual pixels across support and query images, but they suffer from high computational costs due to full dot-product attention and can become unstable when overloaded with dense pixel-wise support features \cite{xu2023self}. A common limitation shared by both approaches is the under utilization of intermediate encoder features during decoding. Most methods rely on shallow or single-scale decoders that do not effectively incorporate mid-level representations from the encoder, missing valuable contextual information. This is particularly problematic in few-shot settings, where richer features are essential for generalizing from limited samples. These challenges highlight a clear gap: the need for a lightweight yet semantically expressive framework that effectively captures both relational understanding and multi-scale context for robust few-shot segmentation.}

	Inspired by recent developments, we aim to develop a straightforward and effective framework to address limitations in FSS methods. A notable approach gaining traction is the Query-based\footnote{For differentiating it from the conventional term "query" frequently employed in FSS, we capitalize "Query" when referring to the query sequence within the Transformer architecture.} Transformer architecture, which has demonstrated versatility across various computer vision tasks, including few-shot learning scenarios \cite{su2024roformer,tian2024survey,luo2023closer,cao2022prototype}. This architecture utilizes learnable Query embeddings derived from support prototypes, enabling nuanced analysis of their relationships within the query feature map. 

	Inspired by previous works, we have designed a novel Transformer-based module, known as the Spatial Transformer Decoder (STD), to enhance the relational understanding between support images and the query image. This module operates concurrently with the multi-scale decoder. The core architecture of our approach is shown in Figure \ref{fig:ch2_adapte_pro}-\subref{fig:1_c}. Within the STD module, we introduce a common strategy: Using the prototype of support images as a Query, while utilizing the features extracted from the query image as both Value and Key embeddings inputted into the Transformer decoder. This formulation allows the Query to effectively focus on the semantic features of the target class within the query image. Furthermore, to reduce the impact of information loss resulting from the abstraction of support images into a feature vector named the 'support prototype,' we integrate global features from the intermediate stages of the encoder, which are fed with the support images, into our decoder. Incorporating these features allows us to leverage features from different stages of the encoder, thereby enriching the decoder's contextual understanding. Additionally, we introduce the Contextual Mask Generation Module (CMGM) to further augment the model's relational understanding (not shown in Figure \ref{fig:ch2_adapte_pro}-\subref{fig:1_c}), operating alongside the STD and enhancing the model's capacity to capture relevant contextual information.
	
		In summary, our contributions include:
		\begin{enumerate}
			\item We propose MSDNet, a novel and lightweight framework for few-shot semantic segmentation, which incorporates a STD. In contrast to conventional designs, our STD uses the support prototype as the Query and the query feature map as the Key and Value in a multi-head cross-attention mechanism, enhancing semantic alignment between support and query features. Despite having only 1.5M learnable parameters, our model achieves competitive performance on standard benchmarks.
			
			\item We introduce a multi-scale decoder architecture that hierarchically refines segmentation masks using progressively integrated mid-level and high-level support features. This approach differs from most prior FSS methods, which commonly rely on shallow or single-scale decoders, and enables more precise mask generation with spatial detail.
			
			\item We develop a novel CMGM, which enhances pixel-wise relational understanding by computing cosine similarities between support and query features. This module provides a semantic prior that guides subsequent processing stages more effectively than traditional feature concatenation or simple prototype averaging.
			
			\item We conduct comprehensive evaluations on the $PASCAL\text{-}5^i$ and $COCO\text{-}20^i$ benchmarks in both 1-shot and 5-shot settings. Our model consistently ranks among the top-performing methods across all folds, confirming its effectiveness and efficiency in a variety of segmentation scenarios.
		\end{enumerate}

	\section{Related Works}
	\subsection{Semantic Segmentation}
	Semantic segmentation, a crucial task in computer vision, involves labeling each pixel in an image with a corresponding class \cite{rizzoli2024source,rezvani2024single,zhou2024cross}. CNNs significantly advanced semantic segmentation by replacing fully connected layers with convolutional layers, enabling the processing of images of various sizes \cite{saber2024efficient,rezvani2025fusionlungnet}. Since then, subsequent advancements have focused on enhancing the receptive field and aggregating long-range context in feature maps. Techniques such as dilated convolutions \cite{yu2015multi}, spatial pyramid pooling \cite{he2015spatial}, and non-local blocks \cite{wang2018non} have been employed to capture contextual information at multiple scales. More recently, Transformer-based backbones, including SegFormer \cite{xie2021segformer}, Segmenter \cite{strudel2021segmenter}, and SETR \cite{zheng2021rethinking}, have been introduced to better capture long-range context in semantic segmentation tasks. Hierarchical architectures like Swin Transformer \cite{liu2021swin} have achieved SOTA performance by computing shifted windows for general-purpose backbones. Additionally, self-supervised pretraining approaches, such as masked image modeling in BEiT \cite{bao2021beit}, have demonstrated competitive results by fine-tuning directly on the semantic segmentation task.
	
	Semantic segmentation tasks typically involve per-pixel classification. as demonstrated by approaches like MaskFormer \cite{cheng2021per} and Mask2Former \cite{cheng2022masked}, which predict binary masks corresponding to individual class labels. Older architectures, such as UNet \cite{ronneberger2015u}, PSPNet \cite{zhao2017pyramid}, and Deeplab \cite{chen2017deeplab,chen2018encoder}, have also significantly contributed to the field by incorporating features like global and local context aggregation and dilated convolutions to increase the receptive field without reducing resolution. Recent studies have sought to improve model performance and contribute to the advancement of semantic segmentation, with notable approaches including CRGNet \cite{xu2022consistency}, and SAM \cite{kirillov2023segment}, among others. While significant progress has been made in understanding and classifying images at the pixel level, further advancements are needed to effectively address the challenge of unseen classes in semantic segmentation.
	
	\begin{figure*}[h]
		\centering
		\includegraphics[width=\linewidth]{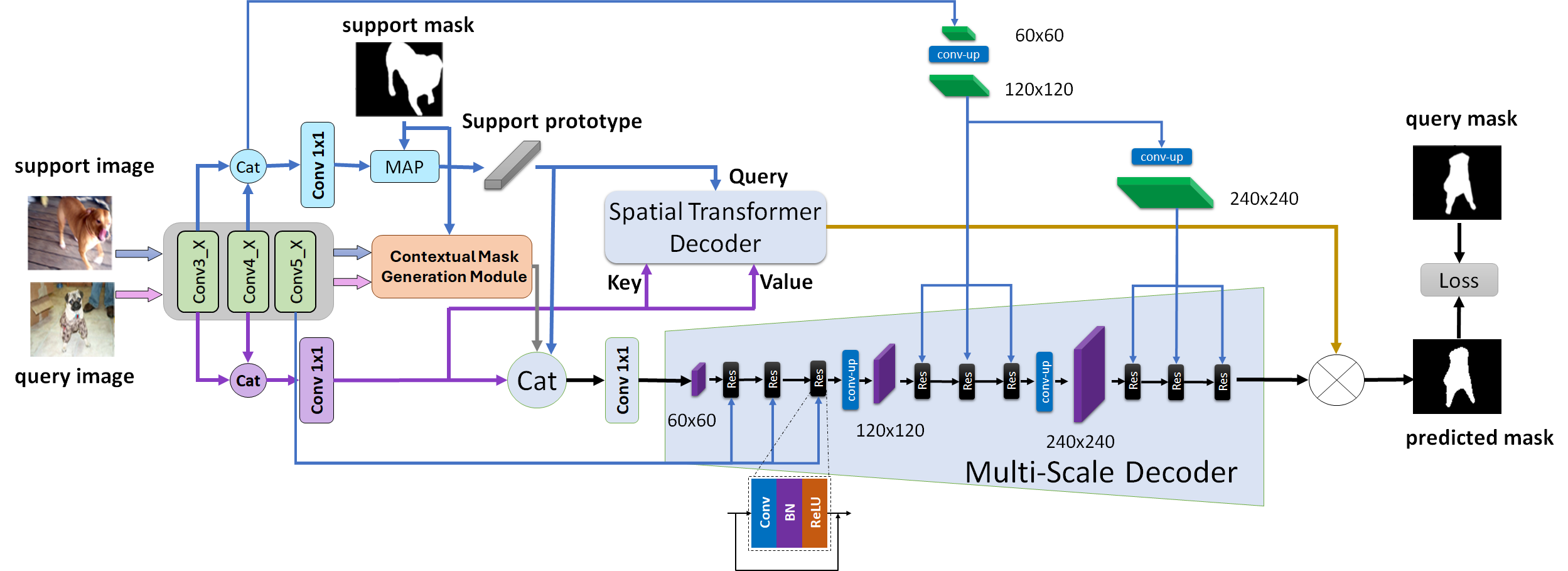}
		
		\caption{the overview of the proposed method}

		\label{fig:overview}
	\end{figure*}
	
	\subsection{Few-Shot Semantic Segmentation}

	FSS is a challenging task in computer vision, wherein the objective is to segment images with limited annotated examples, known as support images. Approaches to FSS can be categorized into various groups based on their primary aims and methodologies employed \cite{li2021adaptive,lu2021simpler,xie2021scale,hong2022cost}. One significant challenge in FSS is addressing the imbalance in details between support and query images. Methods like PGNet \cite{zhang2019pyramid} and PANet \cite{wang2019panet} aim to eliminate inconsistent regions between support and query images by associating each query pixel with relevant parts of the support image or by regularizing the network to ensure its success regardless of the roles of support and query. ASGNet \cite{li2021adaptive}, on the other hand, focuses on finding an adaptive quantity of prototypes and their spatial expanses determined by image content, utilizing a boundary-conscious superpixel algorithm.

	Another critical aspect of FSS is bridging the inter-class gap between base and novel datasets. Approaches like RePRI \cite{boudiaf2021few} and CWT \cite{lu2021simpler} address this gap by fine-tuning over support images or episodically training self-attention blocks to adapt classifier weights during both training and testing phases. Additionally, architectures designed for supervised learning often trouble recognizing objects at different scales in few-shot scenarios. To address this issue, new methods have been developed to allow information exchange between different resolutions \cite{kayabacsi2023elimination,xin2022multilevel}.
	
	Moreover, ensuring the reliability of correlations between support and query images is essential in FSS. Methods like HSNet \cite{min2021hypercorrelation} and CyCTR \cite{zhang2021few} utilize attention mechanisms to filter out erroneous support features and focus on beneficial information. VAT \cite{hong2022cost}, meanwhile, employs a cost aggregation network to aggregate information between query and support features, leveraging a high-dimensional Swin Transformer to impart local context to all pixels.
	
	Overall, the field of FSS is advancing rapidly with innovative methods aimed at enhancing model performance and overcoming challenges in adapting segmentation models to novel classes with limited annotated data. These efforts are driven by the ongoing need to improve the effectiveness and versatility of segmentation models in real-world applications.

	\section{Proposed method}
	
	\subsection{Problem Definition}
	In FSS, the task involves segmenting images belonging to novel classes with limited annotated data. We operate with two datasets, $D_{train}$ and $D_{test}$, each associated with class sets $C_{train}$ and $C_{test}$, respectively. Notably, these class sets are disjoint ($C_{train}\cap C_{test} = \emptyset$), ensuring that there is no overlap between the classes in the training and test datasets. Each training episode consists of a support set $S$ and a query set $Q$, where $S$ includes a set of $k$ support images along with their corresponding binary segmentation masks, while $Q$ contains a single query image. The model is trained to predict the segmentation mask for the query image based on the support set.
	
	Both $D_{train}$ and $D_{test}$ consist of a series of randomly sampled episodes (an episode is defined as a set comprising support images and a query image. During each epoch, we can have many episodes (e.g., 1000 episodes), each containing its own set of support and query images). During training, the model learns to predict the segmentation mask for the query image based on the support set. Similarly, during testing, the model's performance is evaluated on the $D_{test}$ dataset, where it predicts the segmentation mask for query images from the test dataset using the knowledge learned during training.
	
	Overall, the goal of FSS is to develop a model that can accurately segment images from novel classes with only a few annotated samples, demonstrating robust generalization capabilities across different datasets and unseen classes.
	
	\subsection{Overview}

	Given a support set $S=\left \{ I_{s}^{i},M_{s}^{i} \right \}$ and a query image $I_q$, the objective is to generate the binary segmentation mask for $I_q$, identifying the same class as the support examples.  To address this task, we introduce a straightforward yet robust framework, outlined in Figure \ref{fig:overview}. For simplicity, we illustrate a 1-shot setting within the framework, but this can be easily generalized to a 5-shot setting as well. The proposed method comprises several key components, including a shared pretrained backbone, support prototype, Contextual Mask Generation Module (CMGM), a multi-scale decoder, and Spatial Transformer Decoder (STD). These elements collectively contribute to the model's ability to accurately segment objects of interest in the query image based on contextual information provided by the support set. In the following, we'll take a closer look at each component, explaining its role and how it interacts within our framework.
	
	\subsubsection{Backbone}
	
	In our proposed framework, we adopt a modified ResNet architecture, initially pre-trained on the ImageNet dataset, to serve as the backbone for feature extraction from raw input images, ensuring that the size of the output of each block does not reduce below a specified dimension. For instance, like \cite{cao2022prototype}, we define that the output sizes from conv2\_x to conv5\_x are maintained at 60x60 pixels. Specifically, we utilize a ResNet with shared weights between support and query images. This type of ResNet maintains the spatial resolution of feature maps at $60\times60$ pixels from the conv2\_x stage forward, preserving finer details crucial for accurate segmentation. We extract high-level features (conv5\_x), as well as mid-level features (conv3\_x and conv4\_x) from both support and query images using the backbone.
	
	The mid-level features of the support image are denoted as $X_s^{conv3}$ and $X_s^{conv4}$, while the high-level features are denoted as $X_s^{conv5}$. Similarly, for the query image, the mid-level features are represented as $X_q^{conv3}$ and $X_q^{conv4}$, and the high-level features as $X_q^{conv5}$. To integrate mid-level features across different stages, we concatenate the mid-level feature maps from conv3\_x and conv4\_x stages and apply a $1\times1$ convolution layer to yield a merged mid-level feature map, denoted as $X_s^{merged}$. This merging process ensures that the resultant feature map retains essential information from both mid-level stages, enhancing the model's ability to capture diverse contextual information (Equation \ref{merge_s}, Equation \ref{merge_q}). 
	
	\begin{equation}
		X_s^{merged}=C_{1\times1}(Cat(X_s^{conv3},X_s^{conv4}))
		\label{merge_s}
	\end{equation}
	\begin{equation}
		X_q^{merged}=C_{1\times1}(Cat(X_q^{conv3},X_q^{conv4}))
		\label{merge_q}
	\end{equation}
	
	Where $Cat$ denotes concatenation along the channel dimension, and $C_{1\times1}$ denotes the $1\times1$ convolution operation. These equations illustrate the process of merging mid-level features from different stages of the backbone network, resulting in a combined mid-level feature map that retains crucial information from both stages.
	
	The decision to employ this modified ResNet architecture is grounded in its ability to balance computational efficiency with feature representation. By maintaining the feature map size at $60\times60$ pixels, the backbone effectively captures detailed spatial information while avoiding excessive computational overhead. This approach strikes a pragmatic balance between model complexity and segmentation performance, making it well-suited for our few-shot segmentation task, where computational efficiency is paramount.
	
	\subsubsection{Support Prototype}
	
	In our proposed framework, the Support Prototype serves as a condensed representation of the mid-level features extracted from the support example ($X_s^{merged}$). The Support Prototype is obtained by applying a Masked Average Pooling (MAP) operation, which selectively aggregates information based on the support mask. Mathematically, the Support Prototype $P_s$ is defined in Equation \ref{sp}.
	
	\begin{equation}
		P_{s} = F_{pool}(X_{s}^{merged}\odot M_{s})
		\label{sp}
	\end{equation}
	
	Where $F_{pool}$ represents the average pooling operation, and $\odot$ signifies element-wise multiplication (Hadamard product) with the support mask $M_s$. The MAP operation involves computing the average pooling of the masked feature map, focusing solely on regions of interest specified by the support mask. This results in the generation of the Support Prototype, which encapsulates essential semantic information from the support example, facilitating effective few-shot segmentation.
	
	\begin{figure*}
		\centering
		\includegraphics[width=\linewidth]{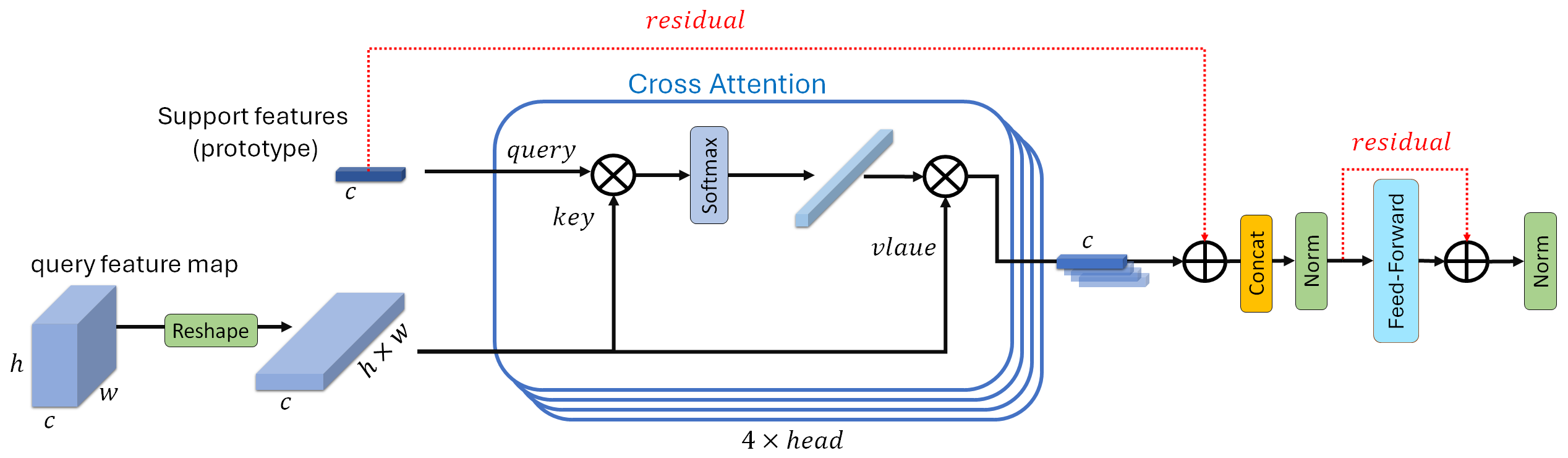}
		
		\caption{Spatial Transformer Decoder}

		\label{fig:STD}
	\end{figure*}
	
	\subsubsection{Contextual Mask Generation Module (CMGM)}
	
	The CMGM is a novel component introduced by our framework, designed to enhance the contextual understanding between support and query images in FSS tasks. At its core, CMGM leverages the feature representations extracted from both the support and query images to generate a contextual mask that encapsulates pixel-wise relations indicative of the target object. This process involves computing the cosine similarity between the query feature vector and the support feature vector. Mathematically, cosine similarity $cos(q,s)$ is calculated as the dot product of the normalized query and support feature vectors. In a five-shot scenario, where there are five support examples, five cosine similarities are computed and subsequently averaged, yielding a novel cosine similarity measure representative of the collective support set. 
	
	The contextual mask produced by CMGM plays a foundational role in guiding the downstream decoder modules. By emphasizing pixel-wise correspondences between the support and query images, CMGM effectively filters the relevant foreground regions. This contextual guidance becomes especially important for the subsequent modules, as it narrows their focus to semantically important regions, allowing them to operate more efficiently and precisely.

	\subsubsection{Multi Scale Decoder}

	The multi scale decoder in our proposed method is a critical component designed to refine the segmentation mask by incorporating features from different resolutions in a hierarchical manner. The decoder consists of three stages, each comprising two residual layers. Input feature map undergoes a sequence of convolutional operations within residual layers to gradually upsample the mask image.
	
	As shown in Figure \ref{fig:overview}, in the first stage of the decoder, the input feature map has a size of $60\times60$ pixels. This stage begins with two residual layers applied to the input feature map. Each residual layer receives input from combination of the previous layer's output and $X_s^{conv5}$. Following these layers, a convolutional operation is employed to upsample the mask image to a resolution of $120\times120$ pixels.
	
	Second stage of the decoder, which operates on a feature map size of $120\times120$ pixels, has two residual layer like the first stage. Each residual layer takes input from combination of the previous layer's output and the merged mid-level features ($X_s^{merged}$) obtained from the support image's encoder.  Since the size of $X_s^{merged}$ remains at $60\times60$ pixels, it is upsampled to $120\times120$ pixel resolution using a convolutional layer. This upsampled feature map, denoted as $X_{s(120\times120)}^{merged}$.
	
	Finally, in the third stage of the decoder, which operates on a feature map size of $240\times240$ pixels, the input to each residual layer comprises the output from the combination of preceding layer and the upsampled $X_s^{merged}$ feature map. in this stage $X_{s(120\times120)}^{merged}$, upsamples to $240\times240$ pixel resolution, denoted as $X_{s(240\times240)}^{merged}$. This upsampled feature map is then integrated with the output from the preceding layer to form the input for subsequent processing.
	
	Notably, one of the distinctive aspects of our multi-scale decoder is the incorporation of mid-level and high-level features from the encoder, like U-Net architecture. Specifically, in each stage of the decoder, the input to the residual layers combines the output from the previous layer with either the $conv5\_x$ features (the output of the last block of the encoder) or the merged mid-level features ($X_s^{merged}$) extracted from the support image's encoder. This fusion of features from different levels of abstraction enhances the decoder's ability to capture both detailed and contextual information essential for accurate segmentation.

	The multi-scale decoder is primarily responsible for spatially refining the segmentation mask by integrating hierarchical feature information. While earlier modules such as CMGM provide a semantic prior for object localization, the multi-scale decoder enhances boundary precision and structural integrity. By leveraging support features at multiple resolutions and combining them through residual connections, the decoder progressively improves the segmentation quality across scales, enabling more detailed and accurate mask reconstruction.

	\subsubsection{Spatial Transformer Decoder (STD)}
	
	In parallel with the multi-scale decoder module, STD plays a pivotal role in refining the final segmentation mask. As illustrated in Figure \ref{fig:STD}, the STD module operates by leveraging multi-head cross-attention, focusing on target objects within the query features to generate semantic-aware dynamic kernels. This process begins by treating the support features as the Query embeddings, while the query features are utilized as the Key and Value embeddings within the STD. Through this strategic integration, the STD module adeptly captures intricate relationships between target objects present in the query features and their corresponding representations in the support features.
	
	The architecture of the STD module employs multi-head cross-attention, rather than a traditional Transformer decoder paradigm. The prototype vector, representing the support features, is integrated as a Query, enriched with learnable positional encodings for heightened spatial context awareness. The query feature map serves as Key and Value embeddings for multi-head cross-attention, enabling comprehensive exploration of their interplay with the support features. Through this multi-head cross-attention process, the STD dynamically generates semantic-aware dynamic kernels crucial for fine-tuning segmentation predictions.
	
	The output of the STD module represents a segmentation mask embedding that captures the semantic information of the target objects within the query features. This embedding is crucial for refining the segmentation results. To integrate this information into the final segmentation output, the segmentation mask embedding is combined with the feature map of the output from the multi-scale decoder using a dot-product operation. This operation efficiently merges the information from both modules, enhancing the overall segmentation accuracy.

	STD serves as a semantic refinement engine that integrates the information distilled by CMGM and complements the spatial reconstruction performed by the multi-scale decoder. By attending to the query features in relation to the contextual support prototypes, STD produces dynamic kernels that capture higher-order dependencies. The resulting semantic-aware embedding is then merged with the output of the multi-scale decoder, allowing the final segmentation prediction to benefit from both semantic precision and spatial detail. This fusion ensures that the strengths of both decoding strategies are harmonized in the final mask generation.

	\subsection{Loss function}
	In our method, we employ the Dice loss function to train our model. This loss function measures the dissimilarity between the predicted segmentation mask $M$ and the corresponding ground truth query mask $M_q$. The Dice loss is formulated in \ref{dice}.
	
	\begin{equation}
		Dice \;  Loss = 1 - \frac{2\times\left | M\bigcap M_q \right |}{\left | M \right |+\left | M_q \right |}
		\label{dice}
	\end{equation}
	
	Where $\left | M\bigcap M_q \right |$ represents the intersection between the predicted and ground truth masks, and $\left | M \right |$ and $\left | M_q \right |$ denote the cardinality of the predicted and ground truth masks, respectively. Minimizing the Dice loss encourages the model to generate segmentation masks that closely match the ground truth masks, leading to more accurate segmentation results during training.

	\section{Experimental Results}

	\subsection{Datasets}
	
	We evaluated our proposed method on two widely used datasets commonly employed in few-shot segmentation tasks: $PASCAL-5^{i}$ \cite{shaban2017one} and $COCO-20^{i}$ \cite{nguyen2019feature}.
	
	\textbf{$\textbf{PASCAL-5}^{\textbf{i}}$ Dataset.}
	The $PASCAL-5^{i}$ dataset, introduced by Shaban et al. \cite{shaban2017one}, is derived from the PASCAL VOC dataset \cite{everingham2010pascal}, and augmented with the SDS \cite{hariharan2011semantic}. The original PASCAL VOC dataset comprises 20 object categories. For $PASCAL-5^{i}$, these 20 categories are evenly divided into 4 subsets, each denoted as $PASCAL-5^{i}$. Consequently, each subset consists of 5 distinct object categories.
	
	\textbf{$\textbf{COCO-20}^{\textbf{i}}$ Dataset.}
	The $COCO-20^{i}$ dataset, introduced by Nguyen et al. \cite{nguyen2019feature}, is derived from MSCOCO dataset \cite{lin2014microsoft}. The $COCO-20^{i}$ dataset includes a total of 80 object categories. Similar to $PASCAL-5^{i}$, these 80 categories are divided into 4 subsets, with each subset denoted as $COCO-20^{i}$. Each subset contains 20 distinct object categories. Notably, $COCO-20^{i}$ presents a greater challenge due to its larger number of categories and images compared to $PASCAL-5^{i}$.
	
	\textbf{Cross-Validation Training.}
	To ensure robust evaluation, we adopted a cross-validation training strategy commonly employed in few-shot segmentation literature. Specifically, we divided each dataset into four subsets. Three subsets were utilized as training sets, while the remaining subset served as the test set for model evaluation. During testing, we randomly selected 1000 support-query pairs from the test set for evaluation.
	
	\subsection{Experimental Setting}
	
	We implemented our proposed method using PyTorch version 1.8.1. For feature extraction, we employed pretrained ResNet-50 and ResNet-101 backbones, which were originally trained on the ImageNet dataset. During training, the parameters of these pretrained models were frozen, and only the newly added modules were trainable. For training on the $COCO-20^{i}$ dataset, we conducted training for each fold over 30 epochs. Conversely, for the $PASCAL-5^{i}$ dataset, training was extended to 60 epochs to ensure optimal convergence. We utilized the Adam optimizer with a fixed learning rate of $10^{-3}$. All input images were resized to $473\times473$ pixels, and the training batch size was set to 32 for the 1-shot setting and 16 for the 5-shot setting. Our training pipeline did not incorporate any data augmentation strategies. After prediction, the binary segmentation masks were resized to match the original dimensions of the input images for evaluation purposes. To ensure robustness and mitigate the effects of randomness, we averaged the results of three trials conducted with different random seeds. All experiments were performed on NVIDIA RTX 4090 GPU.
	
		\begin{table*}[h]
		
		\centering
		\caption{Performance on $PASCAL-5^i$ in terms of mIoU and FB-IoU. Numbers in bold represent the best performance, while underlined values denote the second-best performance.}
		
		\renewcommand{\arraystretch}{1.5}
		\resizebox{2\columnwidth}{!}{
			
			\begin{tabular}{clc|cccccc|cccccc|c}
				\hline
				\multicolumn{1}{c|}{} &
				&
				&
				\multicolumn{6}{c|}{\textbf{1-shot}} &
				\multicolumn{6}{c|}{\textbf{5-shot}} &
				\\
				\multicolumn{1}{c|}{\multirow{-2}{*}{\textbf{Backbone}}} &
				\multirow{-2}{*}{\textbf{Methods}} &
				\multirow{-2}{*}{\textbf{Publication}} &
				fold0 &
				fold1 &
				fold2 &
				fold3 &
				mean &
				FB-IoU &
				fold0 &
				fold1 &
				fold2 &
				fold3 &
				mean &
				FB-IoU &
				\multirow{-2}{*}{\textbf{\begin{tabular}[c]{@{}c@{}}\# learnable\\ params\end{tabular}}} \\ \hline
				&
				PANet \cite{wang2019panet}&
				ICCV19 &
				44.0 &
				57.5 &
				50.8 &
				44.0 &
				49.1 &
				- &
				55.3 &
				67.2 &
				61.3 &
				53.2 &
				59.3 &
				- &
				23.5M \\
				&
				PGNet \cite{zhang2019pyramid}&
				ICCV19 &
				56.0 &
				66.9 &
				50.6 &
				50.4 &
				56.0 &
				69.9 &
				57.7 &
				68.7 &
				52.9 &
				54.6 &
				58.5 &
				70.5 &
				17.2M \\
				&
				PFENet \cite{tian2020prior}&
				TPAMI20 &
				61.7 &
				69.5 &
				55.4 &
				56.3 &
				60.8 &
				73.3 &
				63.1 &
				70.7 &
				55.8 &
				57.9 &
				61.9 &
				73.9 &
				10.3M \\
				&
				PMM \cite{yang2020prototype}&
				ECCV20 &
				52.0 &
				67.5 &
				51.5 &
				49.8 &
				55.2 &
				- &
				55.0 &
				68.2 &
				52.9 &
				51.1 &
				56.8 &
				- &
				- \\
				&
				PPNet \cite{liu2020part}&
				ECCV20 &
				48.6 &
				60.6 &
				55.7 &
				46.5 &
				52.8 &
				69.2 &
				58.9 &
				68.3 &
				66.8 &
				58.0 &
				63.0 &
				75.8 &
				31.5M \\
				&
				RePRI \cite{boudiaf2021few}&
				CVPR21 &
				59.8 &
				68.3 &
				\textbf{62.1} &
				48.5 &
				59.7 &
				- &
				64.6 &
				71.4 &
				\textbf{71.1} &
				59.3 &
				66.6 &
				- &
				- \\
				&
				ASR \cite{liu2021anti}&
				CVPR21 &
				55.2 &
				70.3 &
				53.3 &
				53.6 &
				58.1 &
				- &
				58.3 &
				71.8 &
				56.8 &
				55.7 &
				60.9 &
				- &
				- \\
				&
				SAGNN \cite{xie2021scale}&
				CVPR21 &
				64.7 &
				69.6 &
				57.0 &
				57.2 &
				62.1 &
				73.2 &
				64.9 &
				70.0 &
				57.0 &
				59.3 &
				62.8 &
				73.3 &
				- \\
				&
				HSNet \cite{min2021hypercorrelation}&
				ICCV21 &
				64.3 &
				70.7 &
				60.3 &
				60.5 &
				64 &
				76.7 &
				\underline{70.3} &
				73.2 &
				67.4 &
				\textbf{67.1} &
				\textbf{69.5} &
				80.6 &
				2.5M \\
				&
				CWT \cite{lu2021simpler}&
				ICCV21 &
				56.3 &
				62.0 &
				59.9 &
				47.2 &
				56.4 &
				- &
				61.3 &
				68.5 &
				68.5 &
				56.6 &
				63.7 &
				- &
				- \\
				&
				CyCTR \cite{zhang2021few}&
				NeurIPS21 &
				65.7 &
				71.0 &
				59.5 &
				59.7 &
				64.0 &
				- &
				69.3 &
				73.5 &
				63.8 &
				63.5 &
				67.5 &
				- &
				15.4M \\
				&
				NTRENet \cite{liu2022learning}&
				CVPR22 &
				65.4 &
				\underline{72.3} &
				59.4 &
				59.8 &
				\underline{64.2} &
				77.0 &
				66.2 &
				72.8 &
				61.7 &
				62.2 &
				65.7 &
				78.4 &
				19.9M \\
				&
				ABCNet \cite{wang2023rethinking}&
				CVPR23 &
				62.5 &
				70.8 &
				57.2 &
				58.1 &
				62.2 &
				74.1 &
				64.7 &
				73.0 &
				57.1 &
				59.5 &
				63.6 &
				74.2 &
				- \\
				&
				SRPNet \cite{ding2023self}&
				\begin{tabular}[c]{@{}c@{}}Pattern \\ Recognition23\end{tabular} &
				62.8 &
				69.3 &
				55.8 &
				58.1 &
				61.5 &
				- &
				64.3 &
				70.3 &
				55.1 &
				60.5 &
				62.6 &
				- &
				- \\
				&
				QGPLNet \cite{tang2023query}&
				ACM TOMM23 &
				56.95 &
				68.99 &
				60.1 &
				54.98 &
				60.25 &
				- &
				61.78 &
				70.96 &
				\underline{69.56} &
				58.26 &
				65.14 &
				- &
				- \\
				
				& 
				NSF \cite{lu2023prediction}&
				
				\begin{tabular}[c]{@{}c@{}}IEEE TIP23\end{tabular} &
				51.8 &
				55.4 &
				50.6 &
				36.9 &
				48.7 &
				- &
				59.0 &
				64.0 &
				62.7 &
				48.3 &
				58.5 &
				- &
				- \\
				
				& 
				PCN \cite{lu2023prediction}&
				
				\begin{tabular}[c]{@{}c@{}}IEEE TIP23\end{tabular} &
				47.9 &
				51.2 &
				51.2 &
				41.3 &
				47.9 &
				- &
				53.0 &
				58.0 &
				61.6 &
				51.6 &
				56.0 &
				- &
				- \\
				
				& 
				SRPNet \cite{ding2023self}&
				
				\begin{tabular}[c]{@{}c@{}}Pattern\\ Recognition23\end{tabular} &
				62.8 &
				69.3 &
				55.8 &
				58.1 &
				61.5 &
				- &
				64.3 &
				70.3 &
				55.1 &
				60.5 &
				62.6 &
				- &
				- \\

				&
				DRNet \cite{chang2024drnet}&
				\begin{tabular}[c]{@{}c@{}}IEEE Trans.\\ CSVT24\end{tabular} &
				\underline{66.1} &
				68.8 &
				\underline{61.3} &
				58.2 &
				63.6 &
				76.9 &
				69.2 &
				\underline{73.9} &
				65.4 &
				65.3 &
				68.5 &
				81.6 &
				- \\
				& 
				AFANet \cite{ma2025afanet}&
				
				\begin{tabular}[c]{@{}c@{}}IEEE Trans.\\ Multimedia25\end{tabular} &
				65.7 &
				68.7 &
				60.6 &
				61.5 &
				64.0 &
				- &
				69.0 &
				70.4 &
				61.3 &
				64.0 &
				66.2 &
				- &
				- \\
				& 
				MFIRNet \cite{chen2025mask}&
				
				Neurocomp.25 &
				65.7 &
				69.2 &
				54.5 &
				49.3 &
				59.7 &
				70.4 &
				- &
				- &
				- &
				- &
				- &
				- &
				- \\
				& 
				ESGP \cite{zhang2025efficient}&
				
				\begin{tabular}[c]{@{}c@{}}Pattern\\ Recognition25\end{tabular} &
				63.9 &
				\textbf{72.6} &
				57.1 &
				\underline{61.4} &
				63.8 &
				- &
				- &
				- &
				- &
				- &
				- &
				- &
				- \\
				
				\cline{2-16} 
				\multirow{-23}{*}{ResNet50} &
				\textbf{MSDNet (our)} &
				\textbf{-} &
				\textbf{66.3} &
				71.9 &
				57.2 &
				\textbf{62.0} &
				\textbf{64.3} &
				\textbf{77.1} &
				\textbf{73.2} &
				\textbf{75.4} &
				59.9 &
				\underline{66.3} &
				\underline{68.7} &
				\textbf{82.1} &
				\textbf{1.5M} \\ \hline
				&
				FWB \cite{nguyen2019feature}&
				ICCV19 &
				51.3 &
				64.5 &
				56.7 &
				52.2 &
				56.2 &
				- &
				54.8 &
				67.4 &
				62.2 &
				55.3 &
				59.9 &
				- &
				43.0M \\
				&
				PPNet \cite{liu2020part}&
				ECCV20 &
				52.7 &
				62.8 &
				57.4 &
				47.7 &
				55.2 &
				70.9 &
				60.3 &
				70.0 &
				\underline{69.4} &
				60.7 &
				65.1 &
				77.5 &
				50.5M \\
				&
				DAN \cite{wang2020few}&
				ECCV20 &
				54.7 &
				68.6 &
				57.8 &
				51.6 &
				58.2 &
				71.9 &
				57.9 &
				69.0 &
				60.1 &
				54.9 &
				60.5 &
				72.3 &
				- \\
				&
				PFENet \cite{tian2020prior}&
				TPAMI20 &
				60.5 &
				69.4 &
				54.4 &
				55.9 &
				60.1 &
				72.9 &
				62.8 &
				70.4 &
				54.9 &
				57.6 &
				61.4 &
				73.5 &
				10.3M \\
				&
				RePRI \cite{boudiaf2021few}&
				CVPR21 &
				59.6 &
				68.6 &
				62.2 &
				47.2 &
				59.4 &
				- &
				66.2 &
				71.4 &
				67.0 &
				57.7 &
				65.6 &
				- &
				- \\
				&
				HSNet \cite{min2021hypercorrelation}&
				ICCV21 &
				67.3 &
				72.3 &
				62.0 &
				\textbf{63.1} &
				\textbf{66.2} &
				77.6 &
				71.8 &
				\underline{74.4} &
				67.0 &
				\textbf{68.3} &
				70.4 &
				80.6 &
				2.5M \\
				&
				CWT \cite{lu2021simpler}&
				ICCV21 &
				56.9 &
				65.2 &
				61.2 &
				48.8 &
				58 &
				- &
				62.6 &
				70.2 &
				68.8 &
				57.2 &
				64.7 &
				- &
				- \\
				&
				CyCTR \cite{zhang2021few}&
				NeurIPS21 &
				\textbf{69.3} &
				\underline{72.7} &
				56.5 &
				58.6 &
				64.3 &
				73.0 &
				\underline{73.5} &
				74.0 &
				58.6 &
				60.2 &
				66.6 &
				75.4 &
				15.4M \\
				&
				NTRENet \cite{liu2022learning}&
				CVPR22 &
				65.5 &
				71.8 &
				59.1 &
				58.3 &
				63.7 &
				75.3 &
				67.9 &
				73.2 &
				60.1 &
				66.8 &
				67.0 &
				78.2 &
				19.9M \\
				&
				ABCNet \cite{wang2023rethinking}&
				CVPR23 &
				62.7 &
				70.0 &
				55.1 &
				57.5 &
				61.3 &
				73.7 &
				63.4 &
				71.8 &
				56.4 &
				57.7 &
				62.3 &
				74 &
				- \\
				&
				QGPLNet \cite{tang2023query}&
				ACM TOMM23 &
				59.66 &
				69.77 &
				\textbf{65.15} &
				55.9 &
				62.64 &
				- &
				65.05 &
				72.75 &
				\textbf{71.12} &
				59.85 &
				67.19 &
				- &
				- \\
				
				& 
				NSF \cite{lu2023prediction}&
				
				\begin{tabular}[c]{@{}c@{}}IEEE TIP23\end{tabular} &
				52.6 &
				61.9 &
				58.7 &
				41.5 &
				53.7 &
				- &
				59.9 &
				67.3 &
				65.6 &
				50.4 &
				60.8 &
				- &
				- \\

				&
				DRNet \cite{chang2024drnet}&
				\begin{tabular}[c]{@{}c@{}}IEEE Trans.\\ CSVT24\end{tabular} &
				66.4 &
				70.7 &
				\underline{64.9} &
				59.8 &
				\underline{65.3} &
				\textbf{79.2} &
				69.3 &
				74.1 &
				66.7 &
				66.5 &
				69.2 &
				84.5 &
				- \\
				&
				TBS \cite{park2024task}&
				
				AAAI24 &
				\underline{68.5} &
				72.0 &
				63.8 &
				59.5 &
				65.9 &
				77.7 &
				72.3 &
				74.1 &
				68.4 &
				67.2 &
				\underline{70.5} &
				81.3 &
				- \\
				
				\cline{2-16} 
				\multirow{-16}{*}{ResNet101} &
				\textbf{MSDNet (our)} &
				\textbf{-} &
				67.6 &
				\textbf{72.8} &
				58.2 &
				\underline{60.0} &
				64.7 &
				77.3 &
				\textbf{75.5} &
				\textbf{77.2} &
				62.5 &
				\underline{68.1} &
				\textbf{70.8} &
				\textbf{85.0} &
				\textbf{1.5M} \\ \hline
			\end{tabular}

		}

		\label{tb1}
	\end{table*}

	\begin{table*}[h]
		
		\centering
		\caption{Performance on $COCO-20^i$ in terms of mIoU and FB-IoU. Numbers in bold represent the best performance, while underlined values denote the second-best performance.}
		
		\renewcommand{\arraystretch}{1.5}
		\resizebox{2\columnwidth}{!}{
			
			
			\begin{tabular}{clc|cccccc|cccccc|c}
				\hline
				\multicolumn{1}{c|}{\multirow{2}{*}{\textbf{Backbone}}} &
				\multirow{2}{*}{\textbf{Methods}} &
				\multirow{2}{*}{\textbf{Publication}} &
				\multicolumn{6}{c|}{\textbf{1-shot}} &
				\multicolumn{6}{c|}{\textbf{5-shot}} &
				\multicolumn{1}{c}{\multirow{2}{*}{\textbf{\begin{tabular}[c]{@{}c@{}}\# learnable\\ params\end{tabular}}}} \\
				\multicolumn{1}{c|}{} &
				&
				&
				\multicolumn{1}{c}{fold0} &
				\multicolumn{1}{c}{fold1} &
				\multicolumn{1}{c}{fold2} &
				\multicolumn{1}{c}{fold3} &
				\multicolumn{1}{c}{mean} &
				\multicolumn{1}{c|}{FB-IoU} &
				\multicolumn{1}{c}{fold0} &
				\multicolumn{1}{c}{fold1} &
				\multicolumn{1}{c}{fold2} &
				\multicolumn{1}{c}{fold3} &
				\multicolumn{1}{c}{mean} &
				\multicolumn{1}{c|}{FB-IoU} &
				\multicolumn{1}{c}{} \\ \hline
				\multirow{16}{*}{ResNet50} &
				PPNet \cite{liu2020part} &
				ECCV20 &
				28.1 &
				30.8 &
				29.5 &
				27.7 &
				29.0 &
				- &
				39.0 &
				40.8 &
				37.1 &
				37.3 &
				38.5 &
				- &
				31.5M \\
				&
				PMM \cite{yang2020prototype} &
				ECCV20 &
				29.3 &
				34.8 &
				27.1 &
				27.3 &
				29.6 &
				- &
				33.0 &
				40.6 &
				30.3 &
				33.3 &
				34.3 &
				- &
				- \\
				&
				RPMM \cite{yang2020prototype} &
				ECCV20 &
				\multicolumn{1}{c}{29.5} &
				\multicolumn{1}{c}{36.8} &
				\multicolumn{1}{c}{28.9} &
				\multicolumn{1}{c}{27.0} &
				\multicolumn{1}{c}{30.6} &
				\multicolumn{1}{c|}{-} &
				\multicolumn{1}{c}{33.8} &
				\multicolumn{1}{c}{42.0} &
				\multicolumn{1}{c}{33.0} &
				\multicolumn{1}{c}{33.3} &
				\multicolumn{1}{c}{35.5} &
				\multicolumn{1}{c|}{-} &
				\multicolumn{1}{c}{-} \\
				&
				PFENet \cite{tian2020prior} &
				TPAMI20 &
				36.5 &
				38.6 &
				34.5 &
				33.8 &
				35.8 &
				- &
				36.5 &
				43.3 &
				37.8 &
				38.4 &
				39.0 &
				- &
				10.3M \\
				&
				RePRI \cite{boudiaf2021few} &
				CVPR21 &
				32.0 &
				38.7 &
				32.7 &
				33.1 &
				34.1 &
				- &
				39.3 &
				45.4 &
				39.7 &
				41.8 &
				41.6 &
				- &
				- \\
				&
				HSNet \cite{min2021hypercorrelation} &
				ICCV21 &
				36.3 &
				43.1 &
				38.7 &
				38.7 &
				39.2 &
				68.2 &
				43.3 &
				51.3 &
				48.2 &
				45.0 &
				46.9 &
				70.7 &
				2.5M \\
				&
				CWT \cite{lu2021simpler} &
				ICCV21 &
				32.2 &
				36.0 &
				31.6 &
				31.6 &
				32.9 &
				- &
				40.1 &
				43.8 &
				39.0 &
				42.4 &
				41.3 &
				- &
				- \\
				&
				CyCTR \cite{zhang2021few} &
				NeurIPS21 &
				38.9 &
				43.0 &
				39.6 &
				39.8 &
				40.3 &
				- &
				41.1 &
				48.9 &
				45.2 &
				47.0 &
				45.6 &
				- &
				15.4M \\
				&
				NTRENet \cite{liu2022learning} &
				CVPR22 &
				36.8 &
				42.6 &
				39.9 &
				37.9 &
				39.3 &
				68.5 &
				38.2 &
				44.1 &
				40.4 &
				38.4 &
				40.3 &
				69.2 &
				19.9M \\
				&
				BAM \cite{lang2022learning} &
				CVPR22 &
				\underline{43.4} &
				\textbf{50.6} &
				\textbf{47.5} &
				43.4 &
				\underline{46.2} &
				- &
				\underline{49.3} &
				54.2 &
				51.6 &
				49.6 &
				51.2 &
				- &
				26.7M \\
				&
				DCAMA \cite{shi2022dense} &
				ECCV22 &
				41.9 &
				45.1 &
				44.4 &
				41.7 &
				43.3 &
				69.5 &
				45.9 &
				50.5 &
				50.7 &
				46.0 &
				48.3 &
				71.7 &
				47.7M \\
				&
				ABCNet \cite{wang2023rethinking} &
				CVPR23 &
				36.5 &
				35.7 &
				34.7 &
				31.4 &
				34.6 &
				59.2 &
				40.1 &
				40.1 &
				39.0 &
				35.9 &
				38.8 &
				62.8 &
				- \\
				&
				DRNet \cite{chang2024drnet}&
				\begin{tabular}[c]{@{}c@{}}IEEE Trans.\\ CSVT24\end{tabular} &
				42.1 &
				42.8 &
				42.7 &
				41.3 &
				42.2 &
				68.6 &
				47.7 &
				51.7 &
				47.0 &
				49.3 &
				49.0 &
				71.8 &
				- \\
				&
				QPENet \cite{cong2024query}&
				\begin{tabular}[c]{@{}c@{}}IEEE Trans. \\ Multimedia24\end{tabular} &
				41.5 &
				47.3 &
				40.9 &
				39.4 &
				42.3 &
				67.4 &
				47.3 &
				52.4 &
				44.3 &
				44.9 &
				47.2 &
				69.5 &
				- \\
				&
				PFENet++ \cite{luo2023pfenet++} &
				TPAMI24 &
				\multicolumn{1}{c}{40.9} &
				\multicolumn{1}{c}{46.0} &
				\multicolumn{1}{c}{42.3} &
				\multicolumn{1}{c}{40.1} &
				\multicolumn{1}{c}{42.3} &
				\multicolumn{1}{c|}{65.7} &
				\multicolumn{1}{c}{47.5} &
				\multicolumn{1}{c}{53.3} &
				\multicolumn{1}{c}{47.3} &
				\multicolumn{1}{c}{46.4} &
				\multicolumn{1}{c}{48.6} &
				\multicolumn{1}{c|}{70.3} &
				\multicolumn{1}{c}{-} \\
				&
				DCP \cite{lang2024few}&
				
				\begin{tabular}[c]{@{}c@{}}Int. Jour.\\ Comp. Vision24\end{tabular} &
				43.0 &
				48.6 &
				45.4 &
				44.8 &
				45.5 &
				- &
				47.0 &
				54.7 &
				51.7 &
				\underline{50.0} &
				50.9 &
				- &
				11.3 \\
				&
				PMNet \cite{chen2024pixel}&
				
				WACV24 &
				39.8 &
				41.0 &
				40.1 &
				40.7 &
				40.4 &
				- &
				\textbf{50.1} &
				51.0 &
				50.4 &
				49.6 &
				50.3 &
				- &
				- \\
				&
				RiFeNet \cite{bao2024relevant}&
				
				AAAI24 &
				39.1 &
				47.2 &
				44.6 &
				\underline{45.4} &
				44.1 &
				- &
				44.3 &
				52.4 &
				49.3 &
				48.4 &
				48.6 &
				- &
				- \\
				&
				HSRap
				\cite{luo2025combining}&
				
				\begin{tabular}[c]{@{}c@{}}Exp. System\\ with App.25\end{tabular}&
				43.1 &
				48.5 &
				42.9 &
				41.1 &
				43.8 &
				- &
				49.2 &
				\underline{58.1} &
				\underline{52.9} &
				49.9 &
				\underline{52.5} &
				- &
				- \\

				&
				AFANet \cite{ma2025afanet}&
				
				\begin{tabular}[c]{@{}c@{}}IEEE Trans.\\ Multimedia25\end{tabular} &
				40.2 &
				45.1 &
				44.0 &
				45.1 &
				43.6 &
				- &
				41.0 &
				49.5 &
				43.0 &
				46.9 &
				45.1 &
				- &
				- \\
				\cline{2-16} 
				&
				\textbf{MSDNet (our)} &
				\textbf{-} &
				\multicolumn{1}{c}{\textbf{43.7}} &
				\multicolumn{1}{c}{\underline{49.1}} &
				\multicolumn{1}{c}{\underline{46.9}} &
				\multicolumn{1}{c}{\textbf{46.2}} &
				\multicolumn{1}{c}{\textbf{46.5}} &
				\multicolumn{1}{c|}{\textbf{70.4}} &
				\multicolumn{1}{c}{\textbf{50.1}} &
				\multicolumn{1}{c}{\textbf{58.5}} &
				\multicolumn{1}{c}{\textbf{56.3}} &
				\multicolumn{1}{c}{\textbf{53.1}} &
				\multicolumn{1}{c}{\textbf{54.5}} &
				\multicolumn{1}{c|}{\textbf{74.5}} &
				\multicolumn{1}{c}{\textbf{1.5M}} \\ \hline
				\multirow{12}{*}{ResNet101} &
				FWB \cite{nguyen2019feature} &
				ICCV19 &
				17.0 &
				18.0 &
				21.0 &
				28.9 &
				21.2 &
				- &
				19.1 &
				21.5 &
				23.9 &
				30.1 &
				23.7 &
				- &
				43.0M \\
				&
				PFENet \cite{tian2020prior} &
				TPAMI20 &
				36.8 &
				41.8 &
				38.7 &
				36.7 &
				38.5 &
				63.0 &
				40.4 &
				46.8 &
				43.2 &
				40.5 &
				42.7 &
				65.8 &
				10.3M \\
				&
				HSNet \cite{min2021hypercorrelation} &
				ICCV21 &
				37.2 &
				44.1 &
				42.4 &
				41.3 &
				41.2 &
				69.1 &
				45.9 &
				53.0 &
				51.8 &
				47.1 &
				49.5 &
				72.4 &
				2.5M \\
				&
				CWT \cite{lu2021simpler} &
				ICCV21 &
				30.3 &
				36.6 &
				30.5 &
				32.2 &
				32.4 &
				- &
				38.5 &
				46.7 &
				39.4 &
				43.2 &
				42.0 &
				- &
				- \\
				&
				NTRENet \cite{liu2022learning} &
				CVPR22 &
				38.3 &
				40.4 &
				39.5 &
				38.1 &
				39.1 &
				67.5 &
				42.3 &
				44.4 &
				44.2 &
				41.7 &
				43.2 &
				69.6 &
				19.9M \\
				&
				DCAMA \cite{shi2022dense} &
				ECCV22 &
				41.5 &
				46.2 &
				\underline{45.2} &
				41.3 &
				43.5 &
				69.9 &
				48.0 &
				\underline{58.0} &
				\underline{54.3} &
				47.1 &
				51.9 &
				73.3 &
				47.7M \\
				&
				ABCNet \cite{wang2023rethinking} &
				CVPR23 &
				40.7 &
				45.9 &
				41.6 &
				40.6 &
				42.2 &
				66.7 &
				43.2 &
				50.8 &
				45.8 &
				47.1 &
				46.7 &
				62.8 &
				- \\
				&
				QGPLNet \cite{tang2023query} &
				ACM TOMM23 &
				34.86 &
				40.14 &
				35.68 &
				36.32 &
				36.75 &
				- &
				42.69 &
				48.94 &
				42.98 &
				43.69 &
				44.58 &
				- &
				- \\
				&
				DRNet \cite{chang2024drnet}&
				\begin{tabular}[c]{@{}c@{}}IEEE Trans.\\ CSVT24\end{tabular} &
				43.2 &
				43.9 &
				43.3 &
				\underline{43.9} &
				43.6 &
				69.2 &
				\underline{52.0} &
				54.5 &
				47.9 &
				49.8 &
				51.1 &
				73 &
				- \\
				&
				QPENet \cite{cong2024query}&
				\begin{tabular}[c]{@{}c@{}}IEEE Trans. \\ Multimedia24\end{tabular} &
				39.8 &
				45.4 &
				40.5 &
				40.0 &
				41.4 &
				67.8 &
				47.2 &
				54.9 &
				43.4 &
				45.4 &
				47.7 &
				70.6 &
				- \\
				&
				PFENet++ \cite{luo2023pfenet++} &
				TPAMI24 &
				\multicolumn{1}{c}{42.0} &
				\multicolumn{1}{c}{44.1} &
				\multicolumn{1}{c}{41.0} &
				\multicolumn{1}{c}{39.4} &
				\multicolumn{1}{c}{41.6} &
				\multicolumn{1}{c|}{65.4} &
				\multicolumn{1}{c}{47.3} &
				\multicolumn{1}{c}{55.1} &
				\multicolumn{1}{c}{50.1} &
				\multicolumn{1}{c}{\underline{50.1}} &
				\multicolumn{1}{c}{50.7} &
				\multicolumn{1}{c|}{70.9} &
				\multicolumn{1}{c}{-} \\
				&
				PMNet \cite{chen2024pixel}&
				
				WACV24 &
				\textbf{44.7} &
				44.3 &
				44.0 &
				41.8 &
				43.7 &
				- &
				\textbf{52.6 }&
				53.3 &
				53.5 &
				\underline{52.8} &
				53.1 &
				- &
				- \\
				&
				HSRap
				\cite{luo2025combining}&
				
				\begin{tabular}[c]{@{}c@{}}Exp. System\\ with App.25\end{tabular}&
				42.0 &
				\underline{50.0} &
				43.5 &
				43.8 &
				\underline{44.8} &
				- &
				50.3 &
				\textbf{60.1} &
				53.4 &
				50.9 &
				\underline{53.9} &
				- &
				- \\

				\cline{2-16} 
				&
				\textbf{MSDNet (our)} &
				\textbf{-} &
				\multicolumn{1}{c}{\underline{44.5}} &
				\multicolumn{1}{c}{\textbf{52.5}} &
				\multicolumn{1}{c}{\textbf{48.9}} &
				\multicolumn{1}{c}{\textbf{48.1}} &
				\multicolumn{1}{c}{\textbf{48.5}} &
				\multicolumn{1}{c|}{\textbf{71.3}} &
				\multicolumn{1}{c}{50.4} &
				\multicolumn{1}{c}{\underline{59.9}} &
				\multicolumn{1}{c}{\textbf{57.6}} &
				\multicolumn{1}{c}{\textbf{53.3}} &
				\multicolumn{1}{c}{\textbf{55.3}} &
				\multicolumn{1}{c|}{\textbf{75.1}} &
				\multicolumn{1}{c}{\textbf{1.5M}} \\ \hline
			\end{tabular}

		}

		\label{tb2}
	\end{table*}
	
	\subsection{Evaluation Metrics}
	
	We employ the following evaluation metrics to assess the performance of our proposed method:
	
	\textbf{Mean Intersection over Union (mIoU).}
	mIoU is a widely used metric for evaluating segmentation performance. It calculates the average intersection over union (IoU) across all classes in the target dataset (Equation \ref{miou}).
	
	\begin{equation}
		mIoU=\frac{1}{C}\sum_{i=1}^{C}IoU_{i}
		\label{miou}
	\end{equation} 
	
	Here, $C$ represents the number of classes in the target fold, and $IoU_{i}$ denotes the intersection over union of class $i$.
	
	\textbf{Foreground-Background IoU (FB-IoU).}
	FB-IoU measures the intersection over union specifically for the foreground and background classes. While FB-IoU provides insights into the model's ability to distinguish between foreground and background regions, we primarily focus on mIoU as our main evaluation metric due to its comprehensive assessment of segmentation performance.
	
	\subsection{Comparison with SOTA}
	In this subsection, we compare our proposed method with several SOTA methods on both the $PASCAL-5^i$ and $COCO-20^i$ datasets. We present the results in Table \ref{tb1} and Table \ref{tb2}, respectively, where we report the mIoU and FB-IoU scores under both 1-shot and 5-shot settings, along with the final FB-IoU value. The results of other methods are obtained from their respective original papers.
	
	\textbf{Results on $\textbf{PASCAL-5}^{\textbf{i}}$ Dataset.} As shown in Table \ref{tb1}, our proposed method, utilizing ResNet50 and ResNet101 backbones, consistently surpasses SOTA methods in both 1-shot and 5-shot scenarios across all four folds of the $PASCAL-5^i$ dataset. {Notably, our method achieves competitive performance across all folds, frequently ranking among the top-performing methods in both 1-shot and 5-shot settings.}{}

	\textbf{Results on $\textbf{COCO-20}^{\textbf{i}}$ Dataset.}
	Similarly, Table \ref{tb2} presents the results on the $COCO-20^i$ dataset, where {our proposed method demonstrates strong performance under both ResNet50 and ResNet101 backbones across 1-shot and 5-shot settings. In many folds, our approach achieves the highest or second-highest mIoU scores, reflecting its robustness and efficiency. Additionally, our model achieves competitive mean and FB-IoU scores while maintaining a significantly smaller number of learnable parameters compared to other methods.
	
	Our proposed MSDNet consistently performs well across diverse folds and datasets. In particular, the model shows competitive mIoU scores in fold1 and fold3 on both $PASCAL-5^i$ and $COCO-20^i$, indicating the robustness of our method across varying class distributions. These improvements suggest that MSDNet can generalize well across multiple few-shot segmentation scenarios.
	
	Compared to heavier models such as HSNet \cite{min2021hypercorrelation} and DRNet \cite{chang2024drnet}, MSDNet maintains competitive or superior performance while using significantly fewer parameters. This efficiency stems from our lightweight Transformer-guided decoding strategy and the integration of multi-scale features, which compensates for reduced model size.
	
	In scenarios involving complex object shapes or fine structures, our multi-scale decoder helps refine the mask resolution, especially under the 5-shot setting. However, in some folds with low inter-class variability or where object localization is less ambiguous, larger models with more attention heads (e.g., DCAMA \cite{shi2022dense}) may achieve slightly better results due to their higher modeling capacity.
	
	These findings highlight that MSDNet is especially effective in few-shot settings where efficiency, generalization, and contextual matching are critical, offering a strong balance between accuracy and computational cost.

	}

	
	{
		Our method is designed with computational efficiency in mind. MSDNet contains only 1.5 million learnable parameters, which is significantly fewer than many recent few-shot segmentation models. This lightweight design is particularly beneficial for deployment in real-world scenarios where memory and computational resources are limited.
		
		Although MSDNet integrates two decoding branches—namely the Multi-Scale Decoder and the STD—the architectural design remains computationally tractable. The multi-scale decoder is composed of shallow residual blocks and convolutional upsampling, while the STD is implemented using single cross-attention block rather than a deep transformer stack. This careful design ensures that complexity does not grow excessively, even as the model benefits from richer multi-scale and semantic context.
		
		Furthermore, the low parameter count is achieved without sacrificing segmentation accuracy, as demonstrated in our experimental results. This balance between performance and model efficiency makes MSDNet well-suited for practical applications in environments with constrained compute budgets.
	}
	
		\begin{table*}[!]
		
		\centering
		\caption{Few-shot segmentation performance on cross-dataset task, "$COCO-20^i \rightarrow PASCAL-5^i$", in terms of mIoU, with different backbones (ResNet-50 and ResNet-101). Numbers in bold represent the best performance, while underlined values denote the second-best performance.}
		
		\renewcommand{\arraystretch}{1.5}
		\resizebox{2\columnwidth}{!}{
			
			
			\begin{tabular}{clc|ccccc|ccccc}
				\hline
				\multicolumn{1}{c}{\multirow{2}{*}{\textbf{Backbone}}} &
				\multirow{2}{*}{\textbf{Methods}} &
				\multirow{2}{*}{\textbf{Publication}} &
				\multicolumn{5}{c|}{\textbf{1-shot}} &
				\multicolumn{5}{c}{\textbf{5-shot}}  \\
				\multicolumn{1}{c}{} &
				&
				&
				\multicolumn{1}{c}{fold0} &
				\multicolumn{1}{c}{fold1} &
				\multicolumn{1}{c}{fold2} &
				\multicolumn{1}{c}{fold3} &
				\multicolumn{1}{c|}{mean} &
				\multicolumn{1}{c}{fold0} &
				\multicolumn{1}{c}{fold1} &
				\multicolumn{1}{c}{fold2} &
				\multicolumn{1}{c}{fold3} &
				\multicolumn{1}{c}{mean}  \\ \hline
				\multirow{9}{*}{ResNet50} &
				PFENet \cite{tian2020prior} &
				TPAMI20 &
				43.2 &
				65.1 &
				66.6 &
				69.7 &
				61.1 &
				45.1 &
				66.8 &
				68.5 &
				73.1 &
				63.4  \\
				&
				RePRI \cite{boudiaf2021few} &
				CVPR21 &
				52.2 &
				64.3 &
				64.8 &
				71.6 &
				63.2 &
				56.5 &
				68.2 &
				70.0 &
				76.2 &
				67.7 \\
				&
				HSNet \cite{min2021hypercorrelation} &
				ICCV21 &
				45.4 &
				61.2 &
				63.4 &
				75.9 &
				61.6 &
				56.9 &
				65.9 &
				71.3 &
				80.8 &
				68.7  \\
				&
				VAT \cite{hong2022cost} &
				ECCV22 &
				52.1 &
				64.1 &
				67.4 &
				74.2 &
				64.5 &
				58.5 &
				68.0 &
				72.5 &
				79.9 &
				69.7 \\
				&
				HSNet-HM \cite{liu2022few} &
				ECCV22 &
				43.4&
				68.2&
				\underline{69.4}&
				\textbf{79.9}&
				65.2&
				50.7&
				71.4&
				\underline{73.4}&
				\textbf{83.1}&
				69.7\\
				&
				VAT-HM \cite{liu2022few} &
				ECCV22 &
				68.3&
				64.9&
				67.5&
				\underline{79.8}&
				65.1&
				55.6&
				68.1&
				72.4&
				\underline{82.8}&
				69.7\\
				&
				RTD \cite{wang2022remember} &
				ECCV22 &
				57.4&
				62.2&
				68.0&
				74.8&
				65.6&
				65.7&
				69.7&
				70.8&
				75.0&
				70.1\\
				&
				PMNet \cite{chen2024pixel} &
				WACV24 &
				\underline{68.8}&
				\underline{70.0}&
				65.1&
				62.3&
				\underline{66.6}&
				\textbf{73.9}&
				\underline{74.5}&
				73.3&
				72.1&
				\underline{73.4}\\ \cline{2-13}
				&
				\textbf{MSDNet (our)} &
				\textbf{-} &
				\multicolumn{1}{c}{\textbf{70.7}} &
				\multicolumn{1}{c}{\textbf{73.2}} &
				\multicolumn{1}{c}{\textbf{71.1}} &
				\multicolumn{1}{c}{73.2} &
				\multicolumn{1}{c|}{\textbf{72.1}} &
				\multicolumn{1}{c}{\underline{72.5}} &
				\multicolumn{1}{c}{\textbf{75.0}} &
				\multicolumn{1}{c}{\textbf{73.8}} &
				\multicolumn{1}{c}{75.5} &
				\multicolumn{1}{c}{\textbf{74.2}} \\ \hline 
				\multirow{5}{*}{ResNet101} &
				HSNet \cite{min2021hypercorrelation} &
				ICCV21 &
				47.0 &
				65.2 &
				67.1 &
				\underline{77.1} &
				64.1 &
				57.2 &
				69.5 &
				72.0 &
				\underline{82.4} &
				70.3 \\
				&
				HSNetT-HM \cite{liu2022few} &
				ECCV22 &
				46.7&
				68.6&
				\underline{71.1}&
				\textbf{79.7}&
				66.5&
				53.7&
				70.7&
				75.2&
				\textbf{83.9}&
				70.9\\
				&
				RTD \cite{wang2022remember} &
				ECCV22 &
				59.4&
				64.3&
				70.8&
				72.0&
				66.6&
				67.2&
				72.7&
				72.0&
				78.9&
				72.7\\
				&
				PMNet \cite{chen2024pixel} &
				WACV24 &
				\multicolumn{1}{c}{\underline{71.0}}&
				\multicolumn{1}{c}{\underline{72.3}}&
				\multicolumn{1}{c}{66.6}&
				\multicolumn{1}{c}{63.8}&
				\multicolumn{1}{c|}{\underline{68.4}}&
				\multicolumn{1}{c}{\textbf{75.2}}&
				\multicolumn{1}{c}{\underline{76.3}}&
				\multicolumn{1}{c}{\textbf{77.0}}&
				\multicolumn{1}{c}{72.6}&
				\multicolumn{1}{c}{\underline{75.3}}\\ \cline{2-13}
				&
				\textbf{MSDNet (our)} &
				\textbf{-} &
				\multicolumn{1}{c}{\textbf{71.6}} &
				\multicolumn{1}{c}{\textbf{75.6}} &
				\multicolumn{1}{c}{\textbf{73.0}} &
				\multicolumn{1}{c}{{75.2}} &
				\multicolumn{1}{c|}{\textbf{73.9}} &
				\multicolumn{1}{c}{\underline{71.5}} &
				\multicolumn{1}{c}{\textbf{79.6}} &
				\multicolumn{1}{c}{\underline{76.4}} &
				\multicolumn{1}{c}{{77.9}} &
				\multicolumn{1}{c}{\textbf{76.4}}  \\ \hline
			\end{tabular}

		}

		\label{tb-cross}
	\end{table*}
	
	\subsection{Cross-dataset task}

	In this study, we investigate the cross-domain generalization capabilities of our proposed few-shot segmentation method through rigorous domain shift testing. Specifically, we trained our model on the $COCO-20^i$ dataset and conducted testing on the $PASCAL-5^i$ dataset to evaluate its adaptability across different datasets and domain settings.
	
	The $COCO-20^i$ dataset used in our experiments was modified to exclude classes and associated images that overlap with those present in $PASCAL-5^i$. This adaptation ensured that the training process focused on distinct visual concepts, thereby enhancing the model's exposure to novel classes during testing.
	
	For our experiments, we adopted a cross-dataset evaluation protocol where models trained on each fold of $COCO-20^i$ were repurposed for testing on the entire $PASCAL-5^i$ dataset. Notably, during training, the model was exposed only to specific classes within $COCO-20^i$, ensuring no overlap with the classes present in $PASCAL-5^i$. This setup effectively simulates a scenario where the model encounters novel classes during testing that were not part of its training curriculum.
	
	For instance, in the fold-0 setting, the model was exclusively trained on fold-0 of $COCO-20^i$ and then assessed on the entirety of $PASCAL-5^i$ after filtering out any classes that were encountered during training. This approach tests the model's ability to generalize to new and unseen classes in a different dataset domain.
	
	Our experimental results, as detailed in Table \ref{tb-cross}, demonstrate the superior performance of our proposed method compared to existing SOTA approaches under both 1-shot and 5-shot evaluation scenarios. This underscores the robustness and effectiveness of our few-shot segmentation framework in handling cross-dataset challenges and domain shifts.
	
	\subsection{Ablation Study}
	
	\begin{figure*}[h]
		\centering
		\begin{subfigure}[t]{0.48\linewidth}
			\centering
			\includegraphics[width=\linewidth]{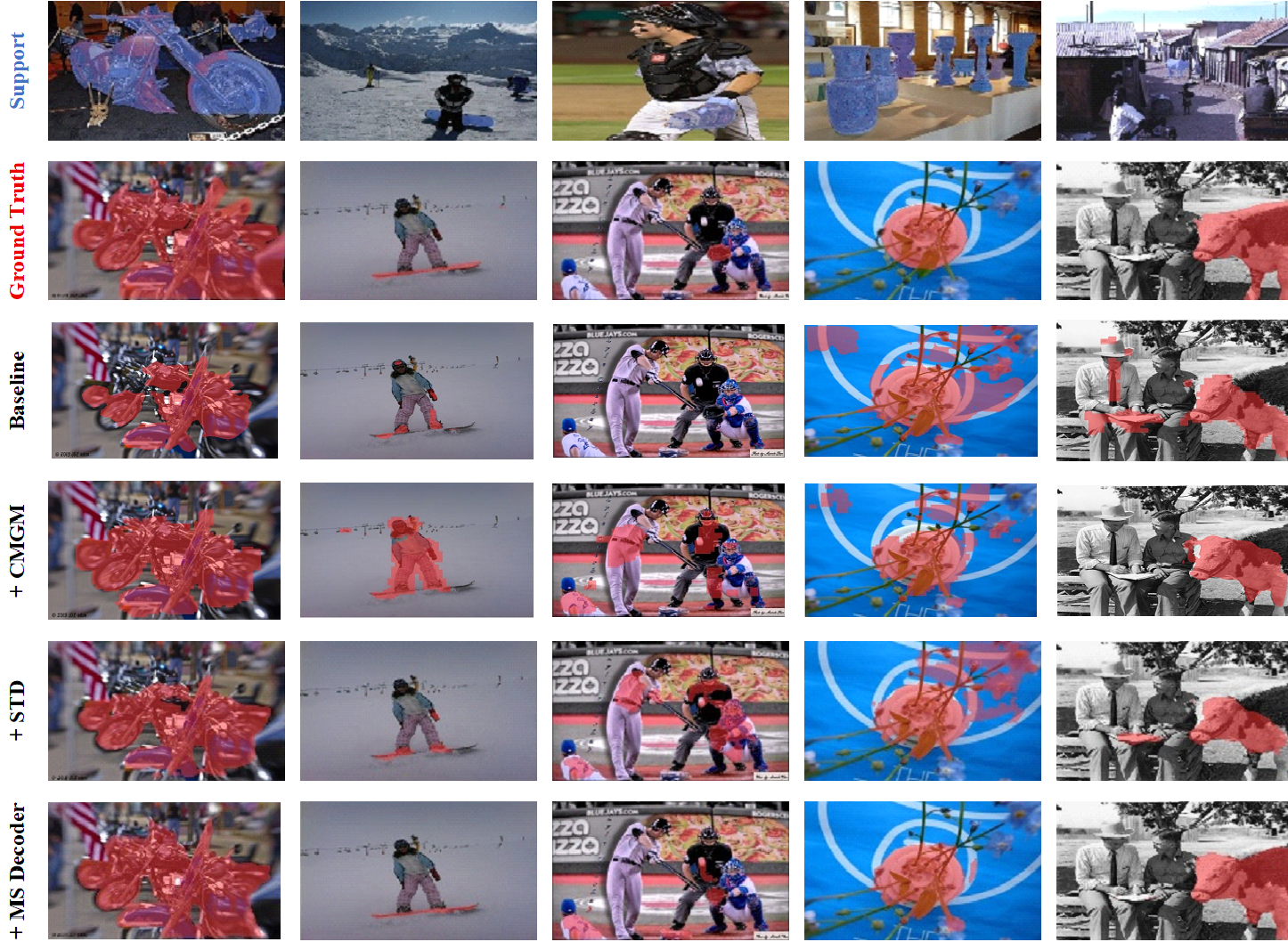}
			\caption{Qualitative comparison of component effects on $COCO\text{-}20^i$ dataset in 1-shot scenario}
			\label{fig:qe_c}
		\end{subfigure}
		\hfill
		\begin{subfigure}[t]{0.48\linewidth}
			\centering
			\includegraphics[width=\linewidth]{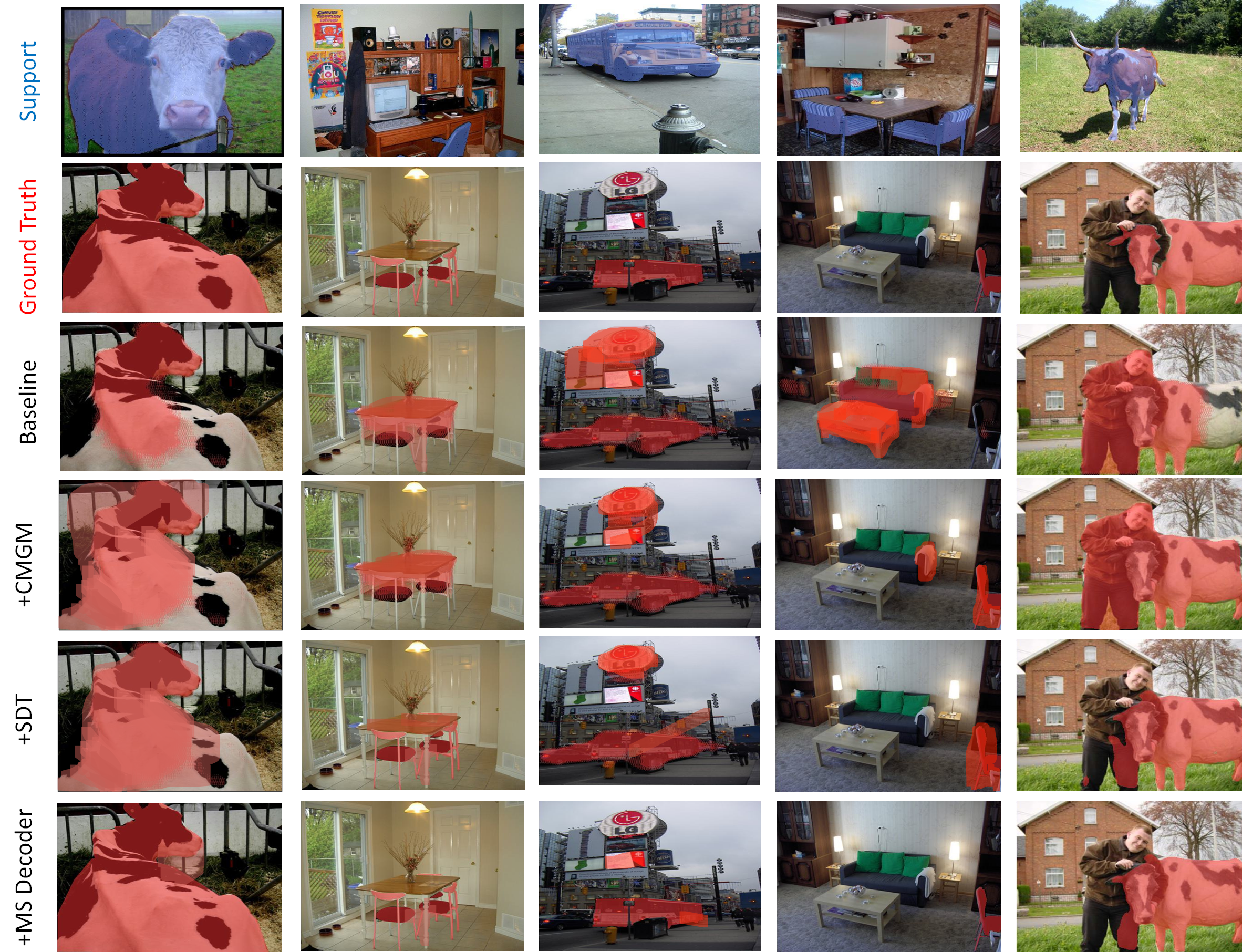}
			\caption{Qualitative comparison of component effects on $Pascal\text{-}5^i$ dataset in 1-shot scenario}
			\label{fig:qe_p}
		\end{subfigure}
		\caption{Qualitative comparison of component effects in 1-shot scenario for (a) $COCO\text{-}20^i$ and (b) $Pascal\text{-}5^i$ datasets.}
		\label{fig:qe}
	\end{figure*}

	To evaluate the contribution of each proposed component, we perform an ablation study on the $COCO\text{-}20^i$ dataset using the ResNet50 backbone under the 1-shot setting. The results are summarized in Table~\ref{tb3}. 
	
	The first row of Table~\ref{tb3} shows the baseline performance, which includes only the backbone and the support prototype mechanism. In the following rows, we incrementally introduce each component—namely, CMGM, STD, and the Multi-Scale Decoder—to analyze their individual and combined effects on segmentation performance.

	\begin{table}[h]
		
		\centering
		\caption{The Impact of Each Component on Segmentation Performance in the $COCO-20^i$ Dataset}
		
		\renewcommand{\arraystretch}{1.5}
		\resizebox{\columnwidth}{!}{
			
			\begin{tabular}{cccc|cccccc}
				\hline
				\multirow{2}{*}{\textbf{Baseline}} &
				\multirow{2}{*}{\textbf{CMGM}} &
				\multirow{2}{*}{\textbf{STD}} &
				\multirow{2}{*}{\begin{tabular}[c]{@{}c@{}}\textbf{Multi Scale} \\ \textbf{Decoder}\end{tabular} } &
				\multicolumn{6}{c}{\textbf{1-shot}} \\
				&     &     &     & \textbf{fold0} & \textbf{fold1} & \textbf{fold2} & \textbf{fold3} & \textbf{mean} & \textbf{FB-IoU} \\ \hline
				$\checkmark$ &     &     &     & 30.1     & 34.2     & 33.4     & 33.8     & 32.9    & 59.7      \\
				$\checkmark$ & $\checkmark$ &     &     & 31.5     & 35.9     & 34.8     & 34.2     & 34.1    & 60.8      \\
				$\checkmark$ &     &$\checkmark$ &      & 34.7     & 40.6     & 34.9     & 37.3     & 36.8    & 63.3      \\
				$\checkmark$ &     &     &$\checkmark$ & 32.1     & 36.8     & 35.2     & 34.6     & 34.7    & 61.2      \\
				$\checkmark$ & $\checkmark$ & $\checkmark$ &     & 43.0     & 45.2     & 43.1     & 41.4     & 43.2    & 67.6      \\
				$\checkmark$ &  $\checkmark$   &      &$\checkmark$ & 36.0    & 40.7     & 36.1     & 37.5     & 37.6    & 63.4      \\
				$\checkmark$ &     & $\checkmark$     &$\checkmark$ & 35.0     & 42.5     & 37.4     & 38.5     & 38.4    & 63.8      \\
				$\checkmark$ & $\checkmark$ & $\checkmark$ & $\checkmark$ & 43.7     & 49.1     & 46.9     & 46.2     & 46.5    & 70.4      \\ \hline
			\end{tabular}

		}
		\label{tb3}
	\end{table}

	As shown in Table \ref{tb3}, each component contributes to an improvement in segmentation performance, with the Multi Scale Decoder showcasing the most substantial impact. The progressive integration of these components results in a notable enhancement in mIoU scores across all folds, underscoring their significance in refining segmentation masks and capturing contextual information effectively.
	
	\begin{table}[h]
		
		\centering
		\caption{The Impact of number of residual blocks in each stage of Multi Scale Decoder on Segmentation Performance in the $COCO-20^i$ Dataset}
		
		\renewcommand{\arraystretch}{1.5}
		\resizebox{\columnwidth}{!}{
			
			\begin{tabular}{ccccccc|c}
				\hline
				\multicolumn{1}{c|}{\multirow{2}{*}{\textbf{\begin{tabular}[c]{@{}c@{}}\# residual\\ blocks\end{tabular}}}} & \multicolumn{6}{c|}{\textbf{1-shot}}     &
				\multicolumn{1}{c}{\multirow{2}{*}{\textbf{\begin{tabular}[c]{@{}c@{}}\# learnable\\ params\end{tabular}}}}      \\
				\multicolumn{1}{c|}{}                                                                                     & \textbf{fold0} & \textbf{fold1} & \textbf{fold2} & \textbf{fold3} & \textbf{mean} & \textbf{FB-IoU} & \\ 
				
				\hline
				1 & 42.4     & 48.2     & 46.0      & 45.1    & 45.4    & 69.4 & 1.0M      \\
				2 & \textbf{43.9}     & 48.9     & 46.7      & 45.5    & 46.2    & 69.7 & 1.2M     \\
				3 & 43.7 & \textbf{49.1} & \textbf{46.9} & \textbf{46.2} & \textbf{46.5} & \textbf{70.4} & 1.5M \\
				4 & 41.9 & 47.4 & 46.4 & 45.8 & 45.4 & 69.5 & 1.7M \\ \hline
			\end{tabular}

		}
		\label{tb4}
	\end{table}

	Furthermore, in Figure~\ref{fig:qe}, we present a qualitative comparison illustrating the effect of progressively adding each proposed component to the baseline model on two benchmark datasets: $COCO\text{-}20^i$ and $Pascal\text{-}5^i$, both under the 1-shot setting. Specifically, Figure~\ref{fig:qe}-\subref{fig:qe_c} shows the results on the $COCO\text{-}20^i$ dataset, while Figure~\ref{fig:qe}-\subref{fig:qe_p} displays the corresponding outcomes on the $Pascal\text{-}5^i$ dataset. As observed in both subfigures, the incorporation of each component consistently leads to noticeable improvements in segmentation quality. In particular, the introduction of the multi-scale decoder contributes significantly to capturing fine-grained details and enhancing object boundaries, thereby demonstrating its effectiveness in improving the overall segmentation performance across diverse image domains.

	\begin{figure}
		\centering
		\includegraphics[width=0.95\linewidth]{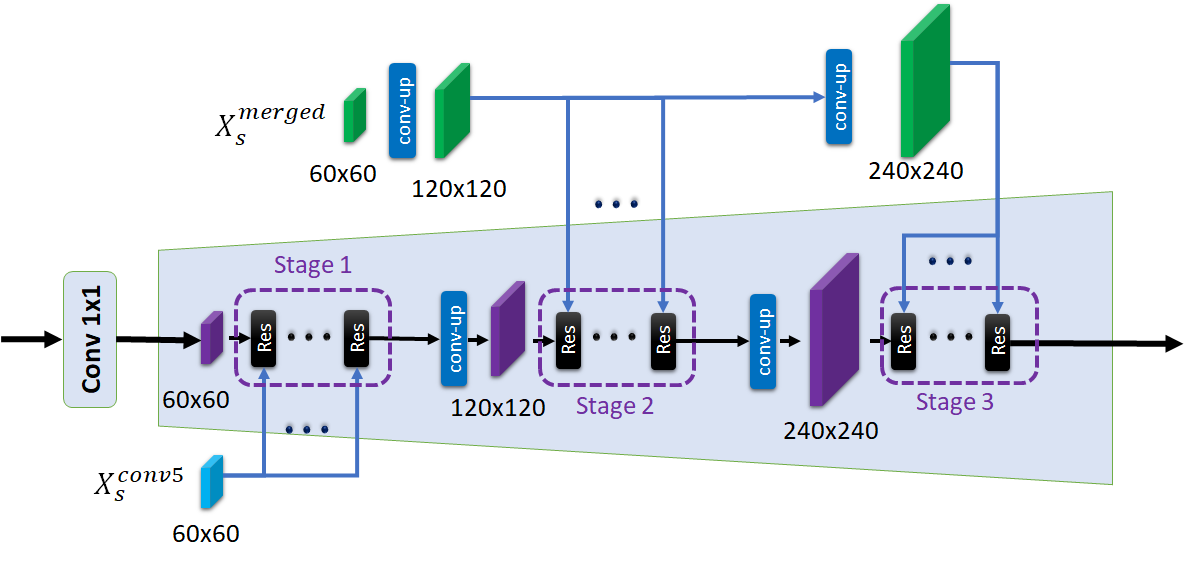}
		
		\caption{The overview of Multi Scale Decoder with different number of residual blocks in each stage (1-4)}

		\label{fig:res_m}
	\end{figure}

	To further explore the influence of the architecture within the multi-scale decoder, we conducted an ablation study varying the number of residual blocks in each stage. Figure \ref{fig:res_m} provides an overview of the Multi-Scale Decoder with different numbers of residual blocks in each stage. The experiment involved evaluating the segmentation performance on the $COCO-20^i$ dataset using the ResNet50 backbone in a 1-shot scenario. As depicted in Table \ref{tb4}, we examined configurations ranging from one to four residual blocks per stage. Interestingly, the results revealed that the optimal segmentation performance was achieved with three residual blocks in each stage. This finding suggests that an appropriate balance in the depth of the decoder architecture plays a crucial role in enhancing segmentation accuracy. Too few blocks may limit the model's capacity to capture intricate features, while an excessive number of blocks could lead to overfitting or computational inefficiency. This experiment also reflects our effort to maintain a lightweight design without compromising performance. The resulting model achieves a low parameter count (1.5M) not through arbitrary reduction, but through deliberate architectural tuning, ensuring both effectiveness and efficiency. Therefore, our results underscore the importance of carefully tuning the architecture parameters to achieve optimal performance in few-shot segmentation tasks.

	\section{Conclusion}
	
	In conclusion, our proposed few-shot segmentation framework, leveraging a combination of components including a shared pretrained backbone, support prototype mechanism, CMGM, STD, and multi-scale decoder, has demonstrated remarkable efficacy in achieving SOTA performance on both $PASCAL-5^i$ and $COCO-20^i$ datasets. Through extensive experimentation and ablation studies, we have highlighted the critical contributions of each component, particularly emphasizing the significant impact of the multi-scale decoder in enhancing segmentation accuracy while maintaining computational efficiency.
	While our method shows strong performance, it is not without limitations. First, the use of a fixed support prototype may oversimplify the representation of intra-class variance in some complex categories. This can lead to reduced accuracy when the support and query images differ significantly in appearance. Second, although our model is lightweight, the presence of dual decoder modules (STD and multi-scale decoder) introduces additional inference time compared to simpler architectures. Lastly, the current architecture is tailored for single-class segmentation per episode; extending it to multi-class few-shot scenarios would require further adaptation and optimization.
	Looking ahead, further investigation into the dynamic adaptation of prototype representations and the exploration of additional attention mechanisms could offer avenues for improving the adaptability and robustness of our method across diverse datasets and scenarios. Additionally, exploring semi-supervised learning paradigms could enhance the generalization capability of our framework, enabling effective segmentation in scenarios with limited labeled data. These avenues for future work hold promise for advancing the effectiveness and applicability of few-shot segmentation methods in real-world scenarios.

	\bibliographystyle{IEEEtran}
	\bibliography{ref1.bib}
	
\end{document}